\documentclass{llncs}
\usepackage{amssymb}
\usepackage{graphicx}

\usepackage[utf8]{inputenc}
\usepackage{times}
\usepackage[colorinlistoftodos]{todonotes}
\usepackage{color}
\usepackage{amsmath}

\usepackage{amsthm}
\usepackage{caption}
\usepackage{subcaption}
\usepackage[hidelinks]{hyperref}
\usepackage{algorithm}
\usepackage{algorithmic}

\title{Effective Multi-Robot Spatial Task Allocation using Model Approximations}
\author{Okan Aşık \and H. Levent Akın}
\institute{Department of Computer Engineering, Boğaziçi University, 34342, Istanbul \\
	\email{\{okan.asik, akin\}@boun.edu.tr}}

\theoremstyle{plain}

\begin{document}

\maketitle

\begin{abstract}
Real-world multi-agent planning problems cannot be solved using decision-theoretic planning methods due to the exponential complexity.
We approximate firefighting in rescue simulation as a spatially distributed task and model with multi-agent Markov decision process.
We use recent approximation methods for spatial task problems to reduce the model complexity.
Our approximations are single-agent, static task, shortest path pruning, dynamic planning horizon, and task clustering.
We create scenarios from RoboCup Rescue Simulation maps and evaluate our methods on these graph worlds.
The results show that our approach is faster and better than comparable methods and has negligible performance loss compared to the optimal policy.
We also show that our method has a similar performance as DCOP methods on example RCRS scenarios.
\end{abstract}

\section{Introduction}
\label{sec:introduction}

Real-world multi-agent planning problems have a high complexity due to the \textit{curse of dimensionality}.
The number of agents also increases the complexity exponentially.
Multi-agent planning can be defined as the coordination of a set of agents to get the highest possible reward from the environment they act in.
Multi-agent planning has different categories, but in this work we consider only the centralized control of cooperative agents.

The spatial task allocation problem (SPATAP) is a subclass of multi-agent planning problems.
In the SPATAP, a group of agents try to do the tasks which are spatially distributed to the environment.
From the multi-agent planning perspective, the SPATAP has two important features; task interdependency, and agent interdependency.
The tasks appear in the environment independently.
Agents move in the environment without affecting each other.
Despite these features, the SPATAP is still a complex multi-agent planning problem and cannot be solved using optimal algorithms as shown by Claes \textit{et. al}~\cite{Claes2015}.

SPATAP is formalized as a Multi-agent Markov Decision Process (MMDP).
The state space is defined by the agent and the task positions.
The agents either move in the environment (such as grid world) or take the action to do the task at the current location.
The reward is defined according to the total task completion.
The complexity of a SPATAP is determined by the state space and the action space.
The optimal MMDP algorithm cannot solve SPATAP problems having non-trivial number of locations, agents, tasks and actions.
There are two basic approaches to solve such complex MMDP problems; using approximate algorithms or approximate models.
In this paper, we propose model approximations which are tailored for the SPATAPs.

The model approximation approach simplifies the given model and finds a solution for the simplified model as a proxy for the actual problem.
The approximations aim to reduce the state space and action space of the actual problem.
Claes \textit{et. al}~\cite{Claes2015} propose a series of approximations for SPATAP planning.
At every time step of the decision process, the method gets the current state of the actual problem and constructs a simpler model using the approximations.
The algorithm calculates a policy for the simple model and the agents act on the actual decision process using the policy.

The approximations proposed by Claes \textit{et. al} are subjective approximation, and phase approximation.
The agent calculates the possible future positions of the agents by the assumption that they are the only agents in the environment.
Then the agent can discount its own future reward according to the possibility of another agent being on that position.
This removes the exponential complexity due to the number of agents in the state space.
The agent assumes that a new task will not appear in the future.
This also reduces the state space complexity due to the future tasks.

We extend the online planning framework of Claes \textit{et. al}~\cite{Claes2015}.
We first cluster tasks based on the distance between the tasks.
Then, we first calculate the best cluster to go using the approximate model.
Then, every agent plans only for the task in the assigned clusters.
In these two levels of the planning, we apply subjective approximation and shortest-path pruning which removes the locations which are not on the shortest path between the agent and the tasks.
We use Value Iteration~\cite{puterman2014markov} algorithm to calculate the best action, but we choose the planning horizon according to the time step required to reach $k$ tasks.

We also generalize the SPATAP model from grid world to graph world where locations are represented by the vertices of a graph.
We define Rescue Spatial Task Allocation Problem (Rescue-SPATAP) as an extension of SPATAP, and solve using SPATAP approximations.
We show that the comparison with the optimal value not as good as pure SPATAP problems, but our method performs better than other algorithms including the SPATAP algorithm~\cite{Claes2015}.
Finally, we apply our SPATAP approximations to RoboCup Rescue Simuation (RCRS) scenarios and have similar performance to Distributed Constraint Optimization (DCOP) methods of RMASBench~\cite{Kleiner2013AAMAS}.

\section{Background}
\label{sec:background}

 \subsection{Multi-agent Markov Decision Process}
 \label{sec:mmdp}
 
 A multi-agent Markov Decision Process (MMDP) is a mathematical formalization for the multi-agent planning in observable, but uncertain action environments.
 
 Multiagent MDP is 5-tuple $\langle D, S, A, T, R \rangle$ where
 \begin{itemize}
 \item $D$ is the set of agents,
 \item $S$ is the finite set of states,
 \item $A$ is the finite set of joint actions ($A_1 \times A_2 \times .. \times A_n$),
 \item $T$ is the transition function which assigns probabilities for transitioning from one state to another given a joint action,
 \item $R$ is the immediate reward function.
 \end{itemize}
 We can solve an MMDP using the standard offline MDP planning algorithms such as value iteration~\cite{puterman2014markov}.
 The value iteration algorithm iteratively improves the estimation of the expected value of a state with the following Bellmann equation:
 
 \begin{align}
  Q(s, a) &= R(s, a) + \gamma \sum_{s^\prime \in S} T(s, a, s^\prime) V(s^\prime) \label{eqn:bellmann_equation}\\
  V(s) &= \max_{a \in A} Q(s, a) \end{align}
  
  $s$ stands for a state, $a$ stands for a joint action and $\gamma$ stands for the discount value to determine how valuable future rewards are.
 

\subsection{Rescue Spatial Task Allocation Problem}
\label{sec:rescue-spatap}

Spatial Task Allocation Problem (SPATAP) is introduced by Claes and \textit{et al.} \cite{Claes2015}.
SPATAP is defined on a location set where a set of different tasks appear on different locations.
Agents have movement actions (to move from one location to another) and also task actions (required to carry out a specific task).
We can think of a grid world where there are two or more cleaning robots.
Cleaning tasks appear at different cells on the grid world.
Agents are supposed to act for cleaning tasks as efficiently as possible.
Although the allocation of agents to their closest tasks would seem to be the optimal, the authors prove that SPATAP is as hard as MMDP.

The original SPATAP formulation models tasks as independent.
Emergence of a task at a location is independent of other locations.
We introduce the Rescue Spatial Task Allocation Problem (Rescue-SPATAP) where tasks are defined as fires and dependent on their neighbors, which makes Rescue-SPATAP harder than SPATAP.
In Rescue-SPATAP, the location of initial tasks/fires are fixed at the start of the process and new tasks only appear based on the vicinity of the current tasks.
In the SPATAP formulation, every task has the same reward, but in Rescue-SPATAP, if the agent extinguishes a fire, it gets a reward proportional to the size of the building.
We show that the online approximations proposed for SPATAPs are also applicable for Rescue-SPATAPs.

The RoboCup Rescue Simulator (RCRS) has four mobile agents: fire brigades, police forces, ambulances, and civilians.
There are also three stationary center agents which provide a communication channel for fire brigades, polices, and ambulances.
There are three types of tasks: rescuing the civilians, firefighting, and removing the blockades on the road.
In this study, we target the firefighting problem, but our approach is also applicable for all RCRS tasks because they can be defined as spatial task allocation problem.
The simulator uses a map of the city.
The map defines buildings and roads.
The simulator creates a disaster scenario by defining fire ignition points.
Since only buildings are flammable, the dynamic tasks emerge as the buildings are catching fire.
The ultimate aim is to develop an algorithm to effectively allocate agents to the buildings which are on fire.

We define a Rescue-SPATAP based on the RCRS.
The problem is defined on a graph world.
The graph world has two types of vertices; buildings and roads as already defined in the RCRS.
In the RCRS, buildings have fire levels: \textit{no fire}, \textit{heating}, \textit{burning}, \textit{burnt}, \textit{extinguished}.
However, to reduce the complexity, we define only two states: \textit{no fire}, and \textit{burning}.
We map the RCRS fire states to the graph world fire states as follows:
\begin{itemize}
\item \textit{no fire} $\leftarrow$ $ \{ $ \textit{no fire}, \textit{burnt}, \textit{extinguished} $ \} $
\item \textit{burning} $\leftarrow $ $ \{ $ \textit{heating}, \textit{burning} $ \} $
\end{itemize}
Since the fire simulator of the RCRS is quite complex to model~\cite{Nussle2005}, we model fire spreading as independent events where the building on fire affects the neighbor buildings' fire state.
Every neighbor building in the vicinity of $d$ meters will add $p$ probability to change the state from \textit{no fire} to \textit{burning}.
The agents move on the graph by choosing the neighbors of the vertex where the agent is on (same as RCRS).
Also, the building the agent is on stays in \textit{no fire} state.
In RCRS, agents extinguish fires based on the size of the building and the water the agent has in its tank.
However, we simplify the fire extinguishing behavior by that an agent extinguishes the fire of the building which the agent is on, regardless of other factors.
The reward is defined as the ratio of the sum of the area of the buildings which are in \textit{no fire} state to the area of all buildings.



	
	




\section{Related Work}

The teams in RoboCup Rescue Agent Simulation (RCRS) generally uses state-based strategies in behavioral agent frameworks~\footnote{\url{http://roborescue.sourceforge.net/blog/2015/08/team-description-papers-tdps/}}.
The teams prefer agent frameworks which enable them to exploit typical scenarios.
These agent frameworks let the teams fine tune their behaviors according to the cases arising over the trial-error periods.
In a recent study, Parker \textit{et al.} report the performance of decentralized coalition formation approach for RCRS~\cite{Parker2015}.
The agents use a greedy algorithm with a utility function which is designed for different tasks.
They compare static and dynamic coalition formation with heterogeneous agents.
Due to the different characteristics of every RCRS scenario, they found that different approaches may work well for different scenarios.

In the literature, the RCRS problem is also modeled as a task allocation problem.
The tasks constitute rescuing a civilian, firefighting, and clearing the blockades.
The tasks are discovered over time and agents do not know all the tasks of the current state.
This distributed dynamic task allocation problem is modeled as the distributed constraint optimization problem (DCOP)~\cite{scerri2005allocating} and solved using state of the art DCOP algorithms such as MaxSum~\cite{Farinelli2008}, and DSA~\cite{Fitzpatrick2003}.
Pujol-Gonzalez \textit{et al.} improve the computational efficiency of MaxSum by introducing Binary MaxSum for RCRS~\cite{Pujol-Gonzalez2015}.
They also introduce a method to integrate team coordination to DCOPs.
The authors show that, by defining coordination variables for police forces and fire brigades, they are able to improve the performance.
Although these approaches have reasonable performances, they require a lot of domain knowledge to design good utility functions with inter-team coordination variables.
Our approach has inherent capacity to represent different agent types without changing the problem definition.

There are also attempts to solve fire task allocation problem with biologically-inspired methods~\cite{dos2011towards}.
They propose a new algorithm, called eXtreme-Ants, where agents are modeled as insects which have response thresholds for tasks that are modeled as stimulus.
They show that the performance of the algorithm is comparable to DCOP methods.

RMASBench is an effort to provide a software repository to easily model RCRS as a DCOP and benchmark the different algorithms~\cite{Kleiner2013AAMAS}.
However, the current implementation requires the full state information of the simulation at every time step and the communication among DCOP agents isolated from RCRS.
This hinders the application of the DCOP methods for RCRS.
Also, modeling RCRS as task allocation problem neglects the dynamic nature of the problem and introduces the issue of designing good utility functions.

\section{Methods}
\label{sec:methods}

We model the firefighting task of the RoboCup Rescue Simulation (RCRS) as a Multi-agent Markov Decision Process(MMDP).
We create an approximate MMDP model with single-agent, static task, shortest path pruning, task clustering, and online planning horizon approximations in our online planning framework.

The online planning framework gets the current state from the simulator (either Rescue-SPATAP simulator or RoboCup Rescue Simulator) and creates a new problem by clustering the near tasks together.
Then, the approximations are applied to the clustered model to have less complex model.
The policy for the approximated model is calculated using the Value Iteration~\cite{puterman2014markov} algorithm.
We calculate the target of every agent by following the policies greedily to assign a cluster to every agent.
Since we assigned a cluster to every agent, the model approximation and policy calculation is carried out considering only the tasks of the assigned clusters.

\subsection{Hierarchical Planning by Task Clustering}
\label{sec:task_clustering}

Before applying any approximations to the actual model of the problem, we create clusters to further reduce the complexity.
Since fires propagate from the initial ignition points, tasks appear as a cluster.
Therefore, we introduce a distance based task clustering algorithm.
The tasks which are closer to each other more than $d$ meters belong to the same cluster.
We assign a cluster for every agent by model approximations and value iteration algorithm.
After every agent is assigned to a cluster, we plan only for the tasks which belongs to the agent's cluster.

\begin{figure}
\centering
\includegraphics[scale=0.35]{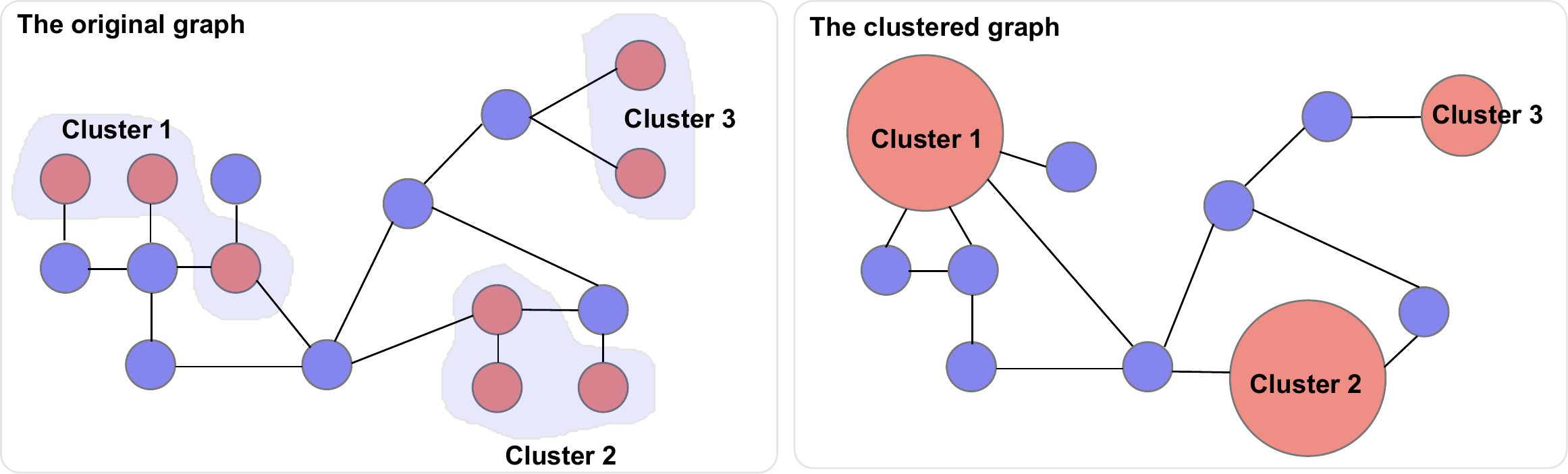}
\caption{The illustration of the task clustering. The initial graph(left) results in a clustered task graph(right).}
\label{fig:graph_clustering}
\end{figure}

To create a cluster, we iterate over all the burning buildings and compare the distance between the building and the buildings in a cluster.
If the building is closer than $d$ meters to one of the buildings which is in a cluster, the building is added to the cluster.
The clustered buildings and their neighbors are removed.
A new building for every cluster is created with the area equal to the sum of the area of the buildings in the cluster.
Neighbors of the clustered buildings are also recreated as the neighbors of the cluster as seen in Figure~\ref{fig:graph_clustering}.

In the SPATAP, the actions are taken according to the calculated value function, but our approach calculates a priority order of tasks for every agent using the depth-first graph traversal algorithm on the value function(taking the agent position as the root).
\subsection{Single-Agent Approximation}
The state space of multi-agent planning has exponential complexity due to the number of agents.
To reduce this complexity, we plan as a single agent by using other agents' positions as an indication of their policies.
We calculate a policy for all agents as if they are the only agent in the environment.
Then, for every agent we calculate the other agents' total effect which is called \textit{presence mass}~\cite{Claes2015}.
\textit{Presence mass} is the probability distribution of other agents' positions on the graph world.

The \textit{presence mass} can be calculated only if we know the policy of the agents.
We calculate a policy for every agent based on their positions on the graph by assuming that they are the only agent on the world.
We use this policy to have an idea about the most desirable action from the perspective of that agent.
To reflect the uncertainty, we use this policy to calculate a Boltzmann distribution over actions for every state.
We define a Boltzmann distribution over the state-action values (the expected cumulative reward when an action is taken in a state).
This distribution defines the probability of choosing an action in a state.

The best response of the agent based on the \textit{presence mass} of other agents can be computed by changing the discount factor($\gamma$) of the Bellman equation (see Equation \ref{eqn:bellmann_equation}).
By changing the discount factor value, we can punish the actions resulting on a position where other agents have high \textit{presence mass}.
Therefore, we discount the future expected reward according to \textit{presence mass} as proposed by ~\cite{Claes2015}:

\begin{align}
 Q_i(s_i,a_i) &= R (s_i, a_i) + \sum_{s_i^{\prime} \in S_i} T (s_i, a_i, s_i^{\prime})  \bigg[ (1 - f_i  \text{ pm}_i(s_i^{\prime})) V_i(s_i^{\prime}) \bigg]
 \label{eqn:vi}\\
 \text{pm}_i(s_i^{\prime}) &= \sum_{j \neq i} Pr (s_j = s_i^{\prime} | s) \nonumber\\
 V_i(s_i) &= \max_{a_i} Q_i(s_i, a_i)
 \nonumber
\end{align}

$Q_i$ denotes the expected total reward for the agent $i$ if it is in state $s_i$, and takes the action $a_i$.
$V_i$ denotes the expected total reward for the agent $i$ from the state $s_i^{\prime}$.
$pm$ defines the \textit{presence mass} of the other agents.
The parameter $f_i$ which is used to scale the future value is calculated as the ratio of maximum reward to the maximum value as suggested by ~\cite{Claes2015}.

\subsection{Static Task Approximation}

We also aim to reduce the exponential complexity due to the fire levels of buildings.
Therefore, we use the approximation proposed by ~\cite{Claes2015} and redefine the state space to include only the buildings that are in \textit{burning} state.
Claes and \textit{et al.}~\cite{Claes2015} propose this approximation for spatially distributed tasks where the occurrence of new tasks are independent.
In the firefighting problem, there is the effect of neighbor buildings on the occurrence of new tasks.
However, the propagation of fire on RCRS is slow such that we can plan considering only the buildings that are on fire without calculating their effects on their neighbors.

The deterministic actions and static task approximations on a graph world for a single agent MDP results in the following Bellman equation:

\begin{equation}
V(s) = \max_{s^\prime \in N(s)} R(s^\prime) + \gamma V(s^\prime)
\label{eqn:static_task}
\end{equation}

$s$ denotes the current state, $s^\prime$ the next state, $N$ the neighbor function, $\gamma$ is the discount factor and $V$ stands for the value function.
Since the actions are deterministic, reward function is only depended on the next state, $s^\prime$, and transition function ($T$) is removed (i.e the action uniquely identifies the next state).
Note that, we changed $(s,a)$ term with $s^\prime$ since actions are deterministic.
Every $(s,a)$ term defines an $s^\prime$ (i.e. every action results in a single next state).
$N$ function defines the set of neighbor vertices of the given vertex (or state).
This recursive equation will be calculated for $h$ times with the initial values $V(s) = 0$ for $h$-horizon planning.



\subsection{The Shortest Path Approximation}
\label{sec:shortest_path}

The tractability of the Bellman equation depends on the state space and the transition function (i.e. neighbors).
If the agent does not move to the vertex which is in \textit{burning} state, the state can only be identified by the position of the agent.
If we remove the neighbors which will not be visited by the agent, we achieve to reduce the state space and also branching factor of the transition function.
The value iteration algorithm propagates rewards from the goal state, in our case this is one of the vertices which is in the \textit{burning} state.


We calculate the shortest path between the agent and the vertices that are in \textit{burning} state, and also all possible \textit{burning} vertex pairs.
Since the actions are deterministic and tasks are static, the optimal policy will result in a movement of the agent on the shortest path from its own position to the one of the tasks.

\subsection{Online Dynamic Planning Horizon}
\label{sec:planning_horizon}

The running time of the value iteration algorithm for the finite horizon problems also depends on the planning horizon which determines the number of iterations of the algorithm.
If we consider Rescue-SPATAP, we should plan according to the current time step of RCRS for the optimal performance.
The RCRS simulation runs for 300 time steps.
For example, if we are in 30\textsuperscript{th} time step, we should construct our approximate model and plan with the value iteration for $300-30 = 270$ time steps.
However, since we already approximated the problem, it might not increase the performance after a certain horizon.
Therefore, we propose to determine the planning horizon based on the reachability of the vertices that are in \textit{burning} state.

\begin{figure}
\centering
\includegraphics[scale=0.35]{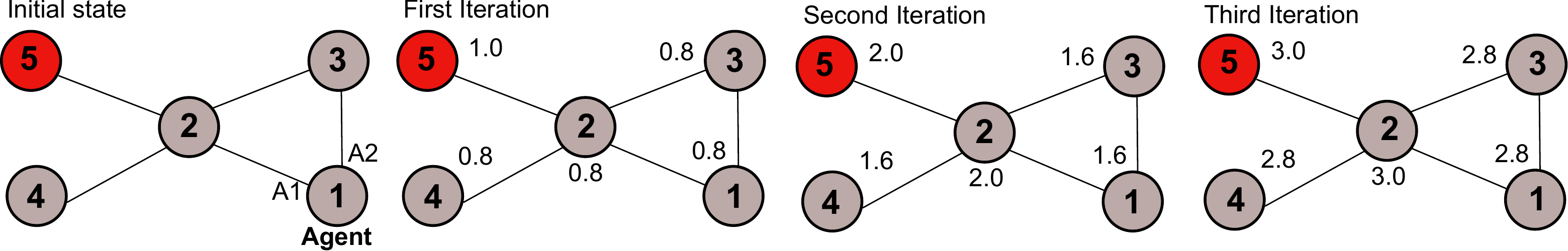}
\caption{Three iterations of the value iteration algorithm on a graph world.}
\label{fig:vi_trace}
\end{figure}

As shown by Claes \textit{et al.}~\cite{Claes2015}, if the agents plan only for the $k$ closest tasks, the algorithm still has reasonable performance.
In the value iteration algorithm, the reward propagates from the vertex that is in \textit{burning} state since the agent gets higher reward when it can change the state of the vertex from \textit{burning} to \textit{no fire}.
For example, we can consider a graph world with the initial state shown in Figure ~\ref{fig:vi_trace}.
There are five vertices where the agent is located on the vertex 1 and the vertex 5 is in \textit{burning} state.
The agent should choose an action based on the values of the vertices 2 and 3.
The three iterations of the value iteration algorithm for the example graph world can be seen in Figure~\ref{fig:vi_trace}.
The values of the vertices correspond to the states where the agent is on that vertex location.
After the first and the second iterations of the algorithm, the agent cannot differentiate between the actions $A1$ and $A2$.
However, after the third iteration of the algorithm, the agent can differentiate two vertices based on the value propagated to the vertex 2 and 3.
In our small graph, three iterations of the algorithm is enough to differentiate two actions, but as the distance of the vertices that are in \textit{burning} state increases, the minimum number of iteration of the algorithm will increase.

To calculate the planning horizon (i.e. the minimum number of iterations), we use the breath first graph traversal algorithm.
We set the vertex of the agent as the root of the graph, and traverse the graph.
When we find $k$ numbers of \textit{burning} vertices, we end the traversal and choose the last level of the tree as our planning horizon and remove the vertices which are not visited from our graph world.

\section{Experiments and Results}


We evaluate the effectiveness of our approach on the sampled graphs from RCRS maps.
All the experiments are implemented using the BURLAP~\cite{burlap} library.

\subsection{Comparison with The Optimal Policy}
\label{sec:exp:optimality}
\begin{figure}
\centering

\begin{subfigure}[b]{0.3\columnwidth}
\frame{\includegraphics[width=\columnwidth]{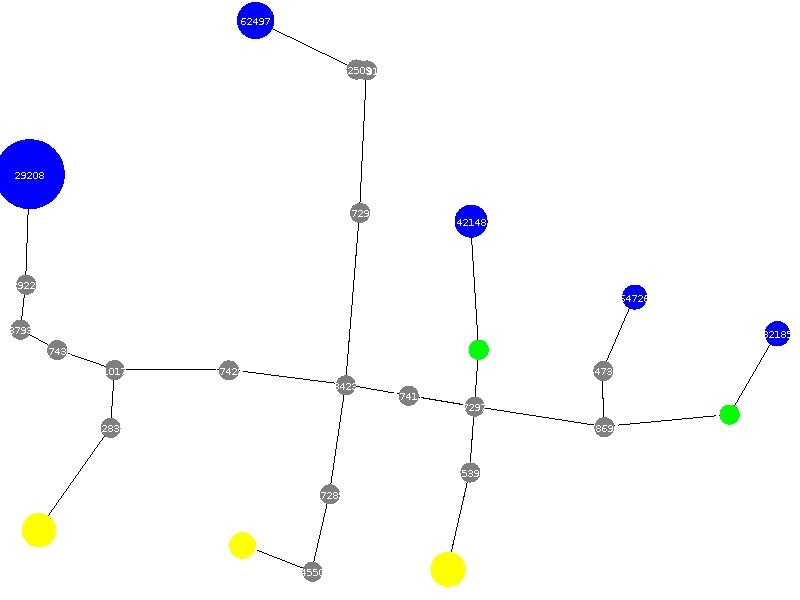}}
\end{subfigure}
~
\begin{subfigure}[b]{0.3\columnwidth}
\frame{\includegraphics[width=\columnwidth]{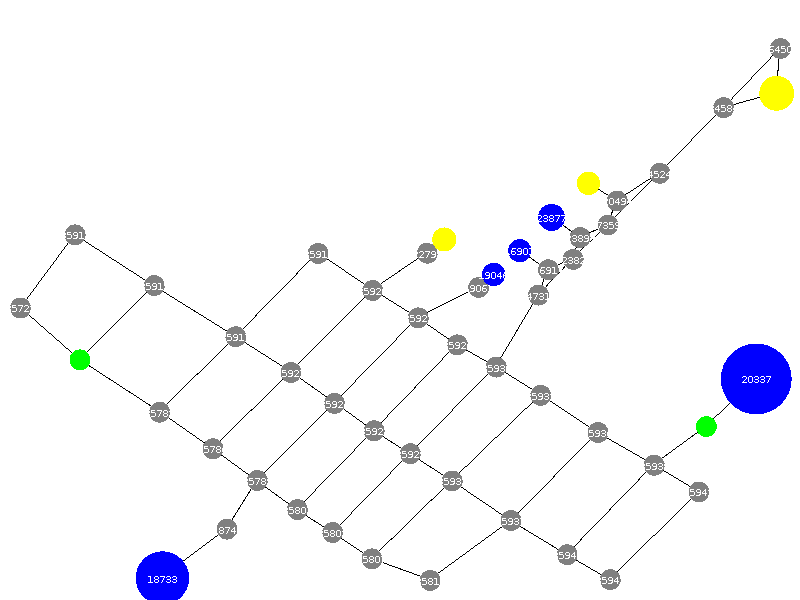}}
\end{subfigure}

\caption{Example sample graphs from Istanbul(left) and Kobe(right) maps having eight buildings (blue) and also road(gray) vertices. The yellow vertices denote the ignition points and the green vertices denotes the position of the agents.}
\label{fig:graph_sampling}
\end{figure}

To measure the feasibility of our approach, we developed a Rescue-SPATAP simulator.
We create 10 random scenarios having 8 buildings from five city graphs of RoboCup Rescue Simulation (RCRS), namely \textit{Istanbul}, \textit{Berlin}, \textit{Eindhoven}, \textit{Joao Pessoa}, and  \textit{Kobe}.
The graph sampling can be seen as the random extraction of districts from a city map (see Figure~\ref{fig:graph_sampling}).
We define three random ignition points and two agents which are positioned on random locations.
In the initial state, ignition buildings are in the \textit{burning} state.
The distance of the building to propagate the fire is $d = 50$ meters and \textit{burning} buildings add $p = 0.05$ probability to their neighbors' fire ignition.
For example, if a building has 2 neighbor \textit{burning} buildings, the probability of changing the state from \textit{no fire} to \textit{burning} will be $(0.05 + 0.05) = 0.1$.
During all the experiments, we use the nearest \textit{burning} building parameter as $k = 3$.
Since we calculate the optimal policy for these simple graph worlds, we model agents' actions as deterministic to reduce the complexity.
This assumption also complies with the RCRS in that the movement noise of agents is almost negligible, if we neglect the congestion of the roads.


We show the average expected reward per time step over a horizon of 20 steps in Table \ref{tbl:optimal_vs_approx}.
We present the results for random agent, single-agent approximation, greedy agent, SPATAP and SPATAP-Ext agents for 50 randomly generated scenarios and 100 samples for every scenario.
Single-agent algorithm plans using the value iteration algorithm as if they were the only agent in the world and act this way.
Greedy agent chooses to go to the closest vertex that is in \textit{burning} state.
Random agents choose random actions.
SPATAP denotes the online approximations proposed by Claes \textit{et al.}~\cite{Claes2015}.
The online approximations proposed in this study as an extension to SPATAP is shown as SPATAP-Ext.
For single-agent and SPATAP algorithms, we coordinate the selection of vertices when two agents are at the same position so that two agents do not choose the same best action.
For greedy algorithm and SPATAP-Ext, we coordinate the choice of the target so that two agents do not go to the same target.
We show that our approach is better than other algorithms.
All the competing algorithms achieve 87\% of optimal average reward, but the SPATAP-Ext achieves 92\%.

%
%
%
%

%
\begin{table}[]
\centering
\caption{The average reward per time step and its percentage with respect to the average optimal value}
	\begin{tabular}{cccccc}
	\hline	
	Random & SA & Greedy & SPATAP & SPATAP-Ext & Optimal \\ \hline
	\begin{tabular}[c]{@{}c@{}}0.365\\$\pm$ 0.154\end{tabular} & \begin{tabular}[c]{@{}c@{}}0.623\\$\pm$ 0.176\end{tabular} & \begin{tabular}[c]{@{}c@{}}0.625\\$\pm$ 0.189\end{tabular} & \begin{tabular}[c]{@{}c@{}}0.621\\$\pm$ 0.170\end{tabular} & \begin{tabular}[c]{@{}c@{}}\textbf{0.654}\\$\pm$ 0.176\end{tabular} & \begin{tabular}[c]{@{}c@{}}0.712 \\$\pm$ 0.147 \end{tabular} \\ \hline
	51.423\% & 87.56\% & 87.75\% & 87.15\% & \textbf{91.88\%} & 100\% \\ \hline
	\end{tabular}
\label{tbl:optimal_vs_approx}
\end{table}



\subsection{Scalability}
\label{sec:exp:scalability}

\begin{figure}
\centering
\begin{subfigure}[b]{0.45\columnwidth}
\includegraphics[scale=0.3]{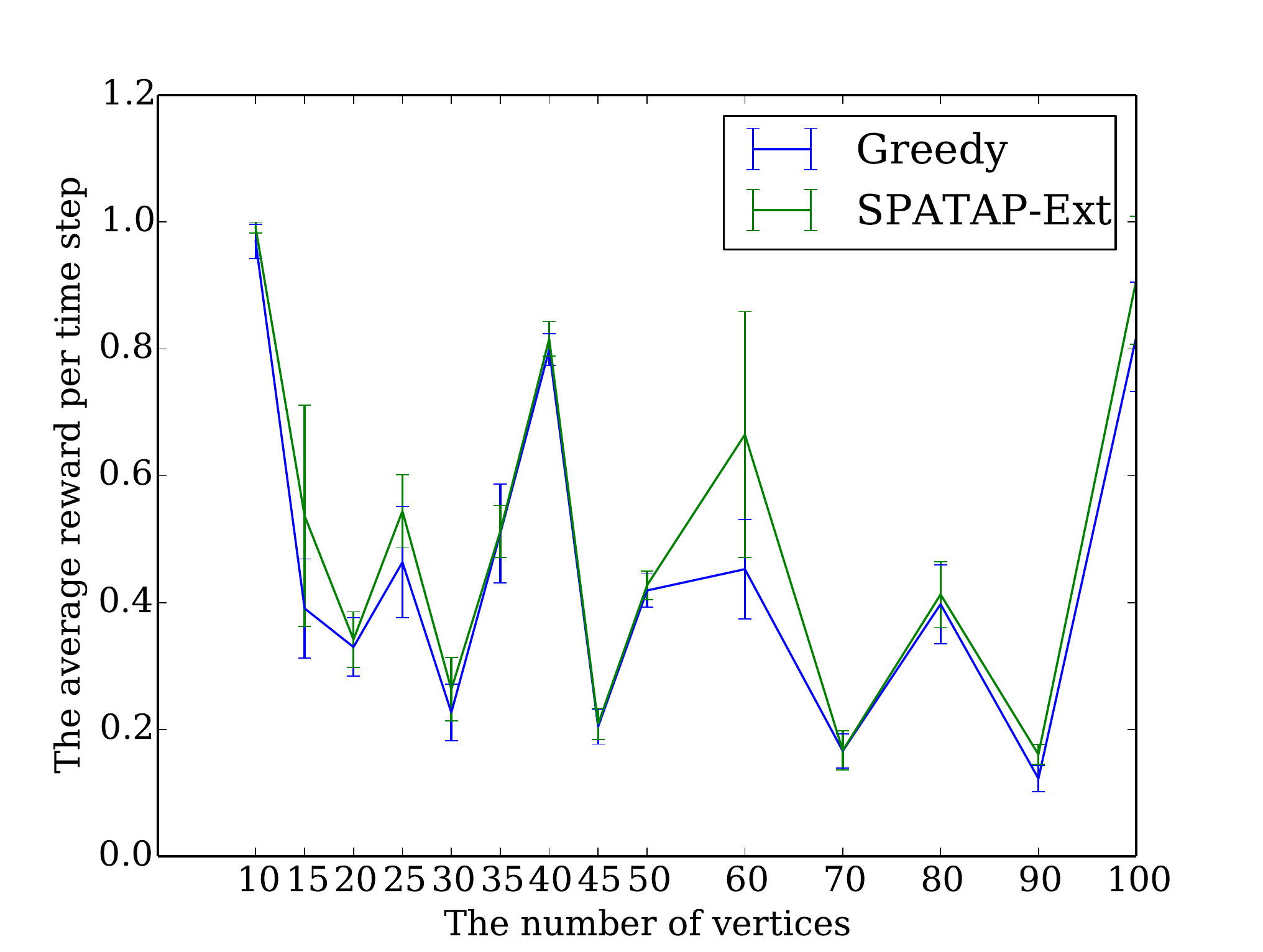}
\caption{The average reward per time step versus the number of vertices}
\label{fig:value_numvertices}
\end{subfigure}
~
\begin{subfigure}[b]{0.45\columnwidth}
\includegraphics[scale=0.3]{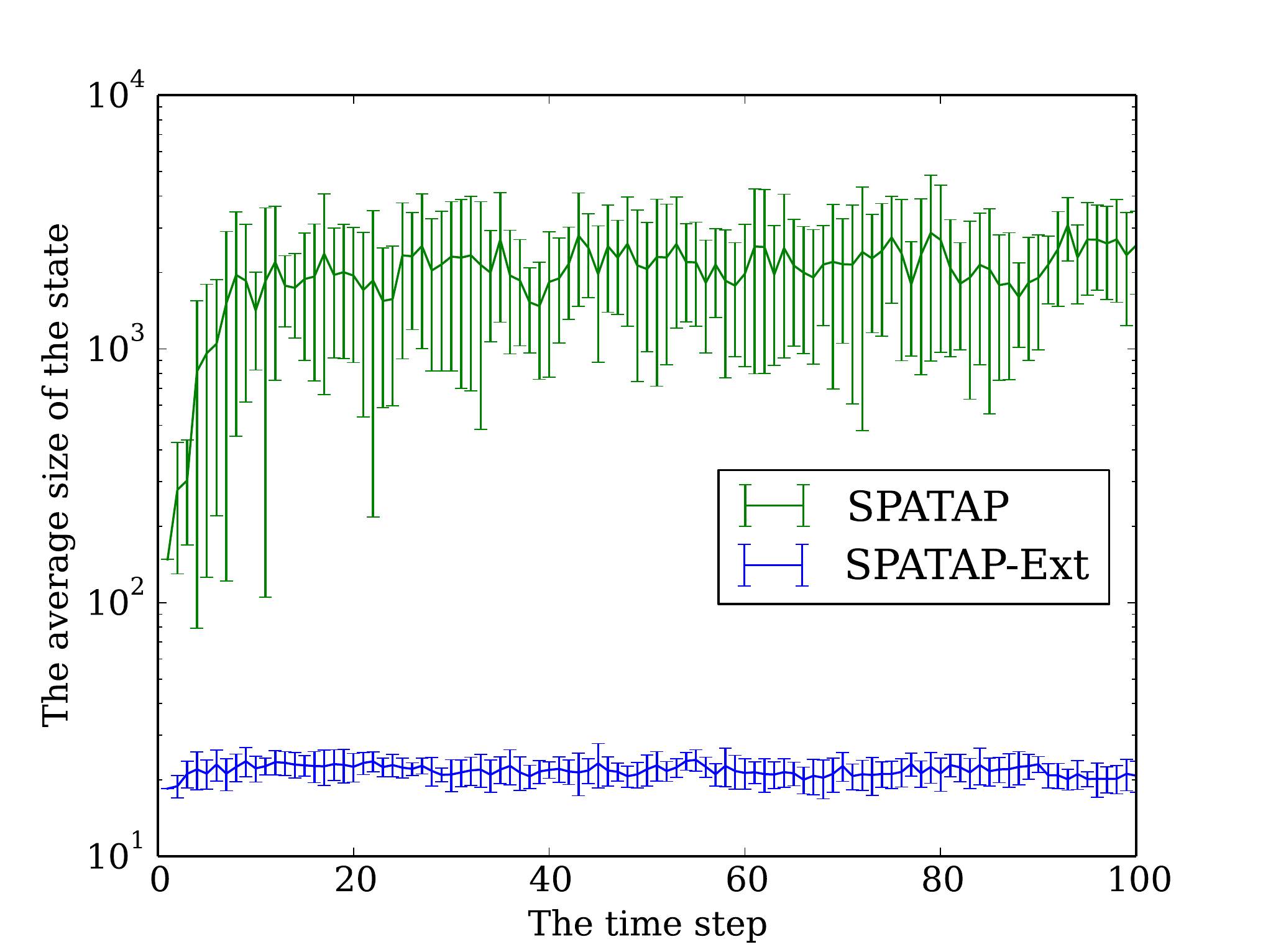}
\caption{The average size of the state versus the time step}
\label{fig:exp_state_space}
\end{subfigure}
\end{figure}

In Figure ~\ref{fig:value_numvertices}, we show the average reward per time step for scenarios having different number of buildings.
These scenarios have five agents and three random ignition points.
The values are averaged over 50 runs and every run is set for 50 horizon.
We can see that SPATAP-Ext performs better or equal compared to Greedy algorithm.
Depending on the ignition points, agents' positions and graph, the difference between two algorithms might increase or decrease.

Due to the approximations, we are able to have linear running time increases when the number of agents or the number buildings increases.
The shortest path pruning and the dynamic planning horizon approximations result in further reduction in the size of the state space.
We show the effect of these extensions in terms of the state space in Figure ~\ref{fig:exp_state_space}.
Our approach reduces the state space by thousandfold compared to SPATAP only approximations.
We use 10 runs of a sample scenario having 10 buildings and 2 agents.

%


\subsection{RoboCup Rescue RMASBench}
\label{sec:exp:rmas}

To benchmark SPATAP-Ext, we created 10 scenarios on the test map (having 37 buildings) of the RCRS.
All of the scenarios~\footnote{Test scenarios: \scriptsize{\url{https://github.com/okanasik/spatial_task_allocation}}} have five to ten ignition points, 8 agents in random positions and 100 horizon~\footnote{An example run can be seen here: \scriptsize{\url{https://youtu.be/nuj8s9aFAlg}}}.
The agents do not act before the $20^{th}$ time step of the simulation to ensure the propagation of the fire.
The score of the RCRS at the end of the simulation is shown in Table~\ref{table:rmasbench}.
Since the fire propagation behavior is not randomized, we report results over a single run.
This score represents the percentage of the damage on the city.
We compare the performance of the SPATAP-Ext with DCOP algorithms (Greedy, DSA, BinaryMaxSum) of the RMASBench~\cite{Kleiner2013AAMAS}.
Although DCOP methods generally perform better than SPATAP-Ext, they have the advantage of well-tuned utility function.
Another important factor affecting the performance of SPATAP-Ext is the assumption that a single agent can extinguish a fire in a single time step irrespective of the size of the building.
This results in the distinct targets for every agent and increases the chances of the propagation of the fire.
In RCRS, if more agents act to extinguish a fire, the faster the fire will be extinguished.
When we analyze the results, we see that even DCOP greedy agent performs better than other methods, this suggest that reflex behavior is more important for such small maps.

\begin{table}[]
\centering
\caption{The comparison of the algorithms for 10 randomly created scenarios on the test map of RCRS.}
\begin{tabular}{@{}ccccc@{}}
\hline
\multicolumn{1}{l}{\textbf{Scenario}} & \textbf{SPATAP-Ext} & \textbf{Greedy} & \textbf{DSA}   & \textbf{BMS}   \\ \hline
1                                     & 0.875               & 0.866           & \textbf{0.878}          & 0.866  \\
2                                     & 0.796               & \textbf{0.814}  & 0.805          & 0.801          \\
3                                     & 0.821               & \textbf{0.844}           & \textbf{0.844} & \textbf{0.844}          \\
4                                     & 0.776               & 0.798           & 0.798          & \textbf{0.810} \\
5                                     & 0.746               & 0.814           & \textbf{0.816}          & \textbf{0.816} \\
6                                     & 0.840               & \textbf{0.868}  & 0.865          & 0.867          \\
7                                     & 0.872               & \textbf{0.885}  & 0.874           & 0.874          \\
8                                     & 0.729               & 0.738           & 0.731          & \textbf{0.745} \\
9                                     & \textbf{0.896}               & 0.890   & 0.890  & 0.890  \\
10                                    & 0.881               & 0.881  & \textbf{0.884}          & 0.881          \\ \hline
\end{tabular}
\label{table:rmasbench}
\end{table}

To increase the performance of SPATAP-Ext agents on RCRS, we also create a set of buildings in unit sizes to enable more agents to extinguish the same building.
Although this improved the performance, we see that agents are more likely to choose the closer buildings.

\section{Conclusion}
\label{sec:conclusion}

We show the application of online approximations for one of the challenging multi-agent planning problems.
Our approach extends SPATAP framework with the introduction of the shortest path pruning, dynamic planning horizon and task clustering approximations for a harder problem Rescue-SPATAP.
We show that our approach is better than Greedy approach and has similar performance to SPATAP, but requires less computation.

As a future work, we plan to extend this framework for heterogeneous agents to model whole RCRS problem.
By introducing partial observability, communication, and decentralized planning, we plan to fully implement online planning framework for RCRS.
We will also reduce the complexity by introducing macro-actions.

\bibliographystyle{splncs03}
\bibliography{asik_robocup16}

\end{document}